\documentclass[11pt]{article}

\usepackage[preprint]{acl}

\usepackage{times}
\usepackage{latexsym}

\usepackage[T1]{fontenc}

\usepackage[utf8]{inputenc}

\usepackage{microtype}

\usepackage{inconsolata}

\usepackage{graphicx}
\usepackage[most]{tcolorbox}
\usepackage{listings}
\usepackage{tabularx}
\usepackage{colortbl}  
\usepackage{array}
\usepackage{makecell}  

\usepackage{amsmath,amssymb,amsthm}
\usepackage{booktabs} 
\usepackage{multirow} 
\usepackage{arydshln} 

\usepackage{xcolor}
\definecolor{green}{rgb}{0.0, 0.5, 0.0} 
\definecolor{red}{rgb}{0.8, 0.0, 0.0} 
\newcommand{\greentext}[1]{\textcolor{green}{#1}} 
\newcommand{\redtext}[1]{\textcolor{red}{#1}} 

\newtcblisting{promptbox}[1]{
    listing only,            
    title={#1},              
    colback=gray!10,            
    colframe=black!70,          
    coltitle=white,             
    fonttitle=\bfseries\sffamily, 
    width=1.0\textwidth,        
    arc=1.5mm,                  
    boxrule=0.5pt,              
    left=4mm, right=4mm,     
    top=2mm, bottom=2mm,     
    listing options={
        basicstyle=\normalfont\small,  
        breaklines=true,         
        breakindent=0pt,         
        columns=fullflexible,    
        keepspaces=true,         
        showstringspaces=false,
    }
}
\newcolumntype{L}{>{\raggedleft\arraybackslash\bfseries}p{2.2cm}}

\usepackage{cleveref}

\title{ReflectRM: Boosting Generative Reward Models via Self-Reflection within a Unified Judgment Framework}

\author{
 \textbf{Kai Qin\textsuperscript{1,2,$\dagger$}},
 \textbf{Liangxin Liu\textsuperscript{2}},
 \textbf{Yu Liang\textsuperscript{2}},
 \textbf{Longzheng Wang\textsuperscript{2}},
 \textbf{Yan Wang\textsuperscript{2}},
\\
 \textbf{Yueyang Zhang\textsuperscript{2}},
 \textbf{Long Xia\textsuperscript{2}},
 \textbf{Zhiyuan Sun\textsuperscript{2}},
 \textbf{Houde Liu\textsuperscript{1,$\ddagger$}},
 \textbf{Daiting Shi\textsuperscript{2,$\ddagger$}},
\\
 \textsuperscript{1}Shenzhen International Graduate School, Tsinghua University
 \\
 \textsuperscript{2}Baidu Inc., Beijing, China
\\
\texttt{tank24@mails.tsinghua.edu.cn}, 
\texttt{\{liuliangxin,liangyu05\}@baidu.com}
\\
\texttt{\{wanglongzheng,wangyan78,zhangyueyang,xialong01\}@baidu.com}
\\ 
\texttt{\{sunzhiyuan01,shidaiting01\}@baidu.com},
\texttt{liu.hd@sz.tsinghua.edu.cn}
\\
}

\begin{document}
\maketitle
{
  \renewcommand{\thefootnote}{\fnsymbol{footnote}}
  \footnotetext[2]{Work done while interning at Baidu.} 
  \footnotetext[3]{Corresponding author.} 
}
\begin{abstract}


Reward Models (RMs) are critical components in the Reinforcement Learning from Human Feedback (RLHF) pipeline, directly determining the alignment quality of Large Language Models (LLMs). Recently, Generative Reward Models (GRMs) have emerged as a superior paradigm, offering higher interpretability and stronger generalization than traditional scalar RMs. However, existing methods for GRMs focus primarily on outcome-level supervision, neglecting analytical process quality, which constrains their potential. To address this, we propose \textbf{ReflectRM}, a novel GRM that leverages self-reflection to assess analytical quality and enhance preference modeling. ReflectRM is trained under a unified generative framework for joint modeling of response preference and analysis preference. During inference, we use its self-reflection capability to identify the most reliable analysis, from which the final preference prediction is derived. Experiments across four benchmarks show that ReflectRM consistently improves performance, achieving an average accuracy gain of +3.7 on Qwen3-4B. Further experiments confirm that response preference and analysis preference are mutually reinforcing. Notably, ReflectRM substantially mitigates positional bias, yielding +10.2 improvement compared with leading GRMs and establishing itself as a more stable evaluator.
Our code is available at \url{https://github.com/yuliangCarmelo/ReflectRM}.

\end{abstract}

\section{Introduction}

The remarkable advancements in LLMs have endowed them with exceptional multi-tasking capabilities, transforming how humans approach complex problem-solving and decision-making \citep{achiam2023gpt, jaech2024openai, liu2024deepseek, team2025kimi, guo2025deepseek, el2025competitive}. To better align these models with human values, RLHF has become a key paradigm, as Supervised Fine-Tuning (SFT) alone often fails to capture the nuanced spectrum of human preferences \citep{christiano2017deep, stiennon2020learning, ouyang2022training, dong2024rlhf, lambert2025reinforcement, li2025omni, seed2025seed1}. Within the RLHF pipeline, RMs play a fundamental role by providing the training signal that ultimately determines the alignment quality of LLMs. Recently, the research focus has shifted to GRMs, a new paradigm that outputs both textual analysis and preference labels  \citep{mahan2024generative, gunjal2025rubrics, jacob2025qa, wang2025gram, ma2025general, liu2025compassverifier, yu2025improve, wu2025meta, ye2025learning}. Compared to traditional scalar RMs, GRMs demonstrate superior evaluative capabilities and better generalization. These advantages make GRMs pivotal for advancing LLMs toward more general proficiency in handling complex and open-ended tasks.

\begin{figure}[!t]
    \centering
    \includegraphics[width=1.0\columnwidth]{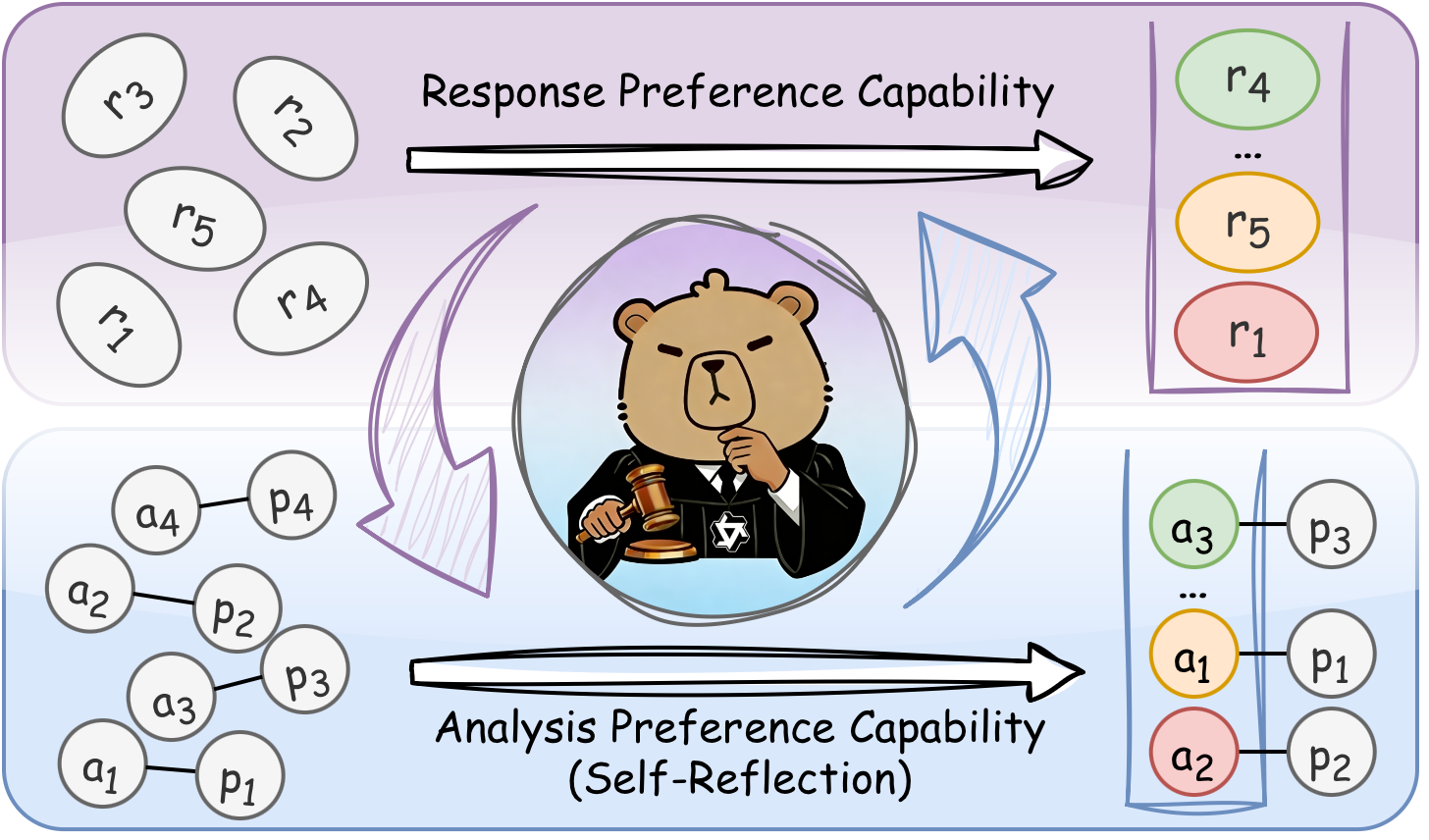} 
    \caption{Mutual reinforcement in ReflectRM. By unifying response preference and analysis preference into a single conditional generative process, ReflectRM internalizes a more robust and reliable evaluative logic.}
    \label{figure:intro}
\end{figure}


Despite rapid advancements in GRMs \citep{zhang2024generative, liu2025inference, chen2025judgelrm, whitehouse2025j1, guo2025reward, chen2025rm}, most existing methods focus primarily on outcome supervision, providing little to no direct supervision over the analytical reasoning process. Representative approaches include J1 \citep{whitehouse2025j1}, which employs online reinforcement learning to generate reasoning for preference judgments, and RM-R1 \citep{chen2025rm}, which combines reasoning distillation via chain-of-rubrics with reward-based optimization. However, both optimize models based primarily on final verdict correctness, without explicit supervision of the analysis process. Although recent work Deepseek-GRM \citep{liu2025inference} has explored trajectory supervision, it formulates process evaluation as an external classification task applied only during inference, requiring auxiliary models and a complex multi-stage pipeline. Consequently, how to provide effective analysis process supervision for GRMs remains an open research question.

In this paper, we propose ReflectRM, a framework that leverages self-reflection to assess analytical quality and improve pairwise preference judgment.
During training, we create process-level preference pairs (Reflection Data) from standard preference tasks by comparing analytical processes leading to correct versus incorrect outcomes.
By integrating standard preference data and reflection data into a unified objective, the model jointly develops two mutually reinforcing capabilities: response preference judgment and analysis preference judgment (self-reflection).
During inference, we adopt a two-stage strategy: the model first generates multiple candidate outputs and selects one with high confidence analysis as an anchor. 
It then employs self-reflection to compare the remaining candidates with this anchor, ultimately selecting the most reliable ones to form the final prediction.

Experimental results across four benchmarks and three model scales demonstrate the consistent effectiveness of ReflectRM. Further analysis reveals a mutually reinforcing relationship between response preference and self-reflection capabilities, which significantly improves the reliability of the model’s core judgments. Moreover, our method effectively mitigates positional bias, leading to substantial gains in positional consistency. In summary, our contributions are as follows:


\begin{itemize}
    \item We propose a unified judgment framework that models both response preference and analysis preference as a single generative task, enabling the unified training of both capabilities.
    \item {ReflectRM} inherently leverages self-reflection to enhance judgments, without requiring auxiliary models or complex training pipelines.
    \item Experimental results across four benchmarks and three different models demonstrate the effectiveness of ReflectRM. 
    \item Further analysis reveals that response preference and self-reflection are mutually reinforcing capabilities, leading to substantial mitigation of positional bias.
\end{itemize}

\begin{figure*}[!t]
    \centering
    \includegraphics[width=1.0\textwidth]{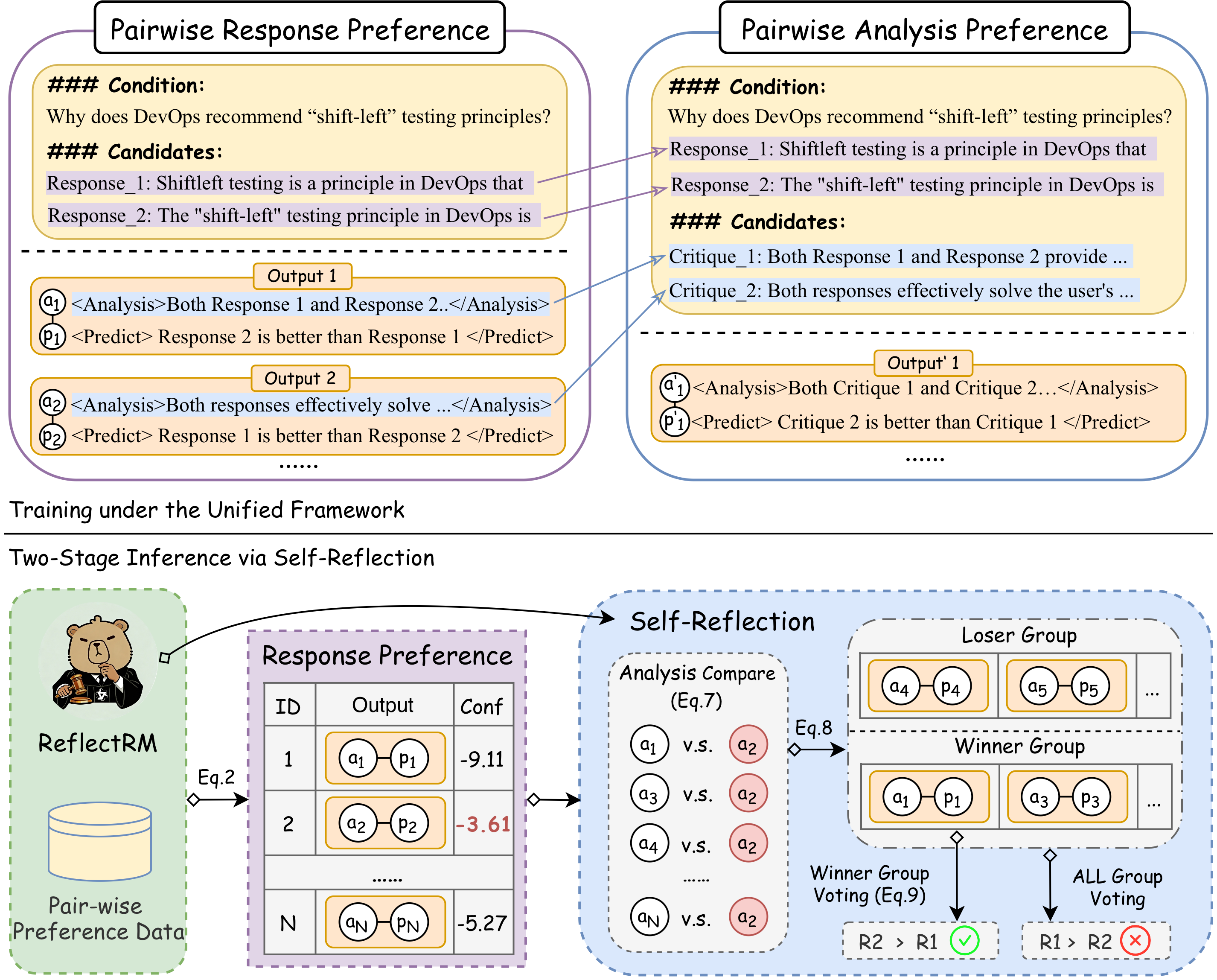} 
    \caption{Overview of the ReflectRM method. \textbf{(Top)} Our \textit{Unified Judgment Framework}, which models response preference and analysis preference as a single conditional generative process. \textbf{(Bottom)} The \textit{Two-Stage Inference Strategy} that leverages the model’s self-reflection capability to identify and aggregate reliable analytical traces, yielding more robust and reliable final judgments.}
    \label{figure:main_pipeline}
\end{figure*}

\section{Related Work}

\paragraph{Generative Reward Models}

Recent studies indicate a shift in research focus from traditional scalar RMs toward GRMs. GRMs are regarded as a promising direction in reward modeling, owing to their superior interpretability and generalization capabilities. For instance, DeepSeek-GRM~\cite{liu2025inference}  introduces a pointwise generative reward model trained via reinforcement learning to generate detailed critiques and self-derived evaluation rubrics, thereby enabling more flexible and task-agnostic scoring than scalar RMs. Following this generative perspective, RM-R1~\cite{chen2025rm} proposes a two-stage pipeline that first distills high-quality reasoning traces and then applies reinforcement learning with verifiable reward signals. Similarly, the Reward Reasoning Model~\cite{guo2025reward} extends GRMs by incorporating explicit reasoning steps before final judgment, even without annotated reasoning traces. Despite these rapid advances, current GRM methods remain limited: they primarily provide general rewards based on outcome-level supervision, overlooking the optimization of the analytical process, which in turn constrains further performance improvements.

\paragraph{Self Reflection in LLMs}

Self-reflection \cite{renze2024self, liu2024self} refers to the ability of LLMs to evaluate, critique, and iteratively improve their own outputs. This capability allows models to analyze initial responses in a structured manner, refine them, and ultimately generate answers of higher quality and accuracy. Demonstrating its effectiveness, the TASTE framework \cite{wang2024taste} achieves significant performance gains in translation by first assessing the quality of its initial output and then refining it based on this assessment. Similarly, SelectIT \cite{liu2024selectit} utilizes self-reflection for high-quality data selection without relying on external models. A further advancement is Progressive Self-Reflection (PSR) \cite{phan2025think}, a novel inference-time technique that empowers LLMs to dynamically self-monitor and correct their outputs. While these methods have proven successful for LLMs, their application to GRMs remains unexplored.

%


\paragraph{Uniqueness of ReflectRM}
ReflectRM distinguishes itself by unifying response and analytical preferences into a single generative process. Unlike prior GRMs that treat reasoning as an unconstrained byproduct of outcome optimization, ReflectRM explicitly supervises the analytical process within a preference-based architecture. This unified framework ensures that both capabilities stem from a shared evaluative logic, making them mutually reinforcing rather than competing. Furthermore, by eliminating complex multi-stage training pipelines and auxiliary classifiers, ReflectRM offers a streamlined and scalable method for advancing generative reward modeling.

\section{ReflectRM}
\label{sec:reflectrm}


Most current reward models are supervised solely based on the correctness of final outcomes while neglecting the explicit supervision of the analytical process. Self-reflection refers to the model's capability to evaluate the analysis and refine its own reasoning, which has proven effective in enhancing output quality. However, this capability has rarely been integrated into the training of reward models. To bridge this gap, we propose \textbf{ReflectRM}, a unified framework that incorporates both process and outcome supervision into reward modeling.

\subsection{A Unified Judgment Framework}
\label{ssec:unified_framework}


In this section, we propose a novel framework that unifies response preference and analysis preference into a single, cohesive generative structure. 

Our core idea is to formulate the supervision of the analytical process as a preference task, and frame all preference tasks as a \textit{conditional generative process}. Given a specific condition $\phi$ and candidates $\delta$, this process aims to generate a comprehensive judgment text, comprising a textual analysis $a$ and a prediction $p$. This process can be formally expressed as:

\begin{equation}
\underbrace{(a, p)}_{\text{O}} \sim f_{\theta}(\ \cdot \mid \underbrace{\phi; \delta}_{\text{I}})
\label{eq:unified_perspective}
\end{equation}
where $f_\theta$ represents the LLM parameterized by $\theta$, while $I = (\phi; \delta)$ and $O = (a, p)$ denote the input and output of the model.
By unifying different preference tasks under this unified generative formulation, ReflectRM acquires a more fundamental and generalized preference judgment capability. This formulation naturally induces two distinct but mutually reinforcing capabilities simply by varying the conditions $\phi$ and candidates $\delta$.

\paragraph{Capability 1: Pairwise Response Preference}

The first capability derived from our unified framework is the preference of responses, which serves as the core reward modeling capability. This is instantiated from \Cref{eq:unified_perspective} by setting the condition $\phi$ to the user's query $q$ and the candidates $\delta$ to the two responses $(r_1, r_2)$:
\begin{equation}
(a, p) \sim f_\theta(\ \cdot \mid q; r_1, r_2)
\label{eq:GRM}
\end{equation}
As illustrated in the "Pairwise Response Preference" box in \Cref{figure:main_pipeline}, this directs the model to perform the primary task of judging which response is better while providing a supporting analysis.

\paragraph{Capability 2: Pairwise Analysis Preference (Self-Reflection)}
%
Our unified framework also introduces pairwise analysis preference, or \textbf{Self-Reflection}, which enables the model to evaluate the quality of its own analytical processes.
To achieve this, the condition $\phi$ in \Cref{eq:unified_perspective} is expanded to the triplet $(q, r_1, r_2)$ to provide the full context, while the candidates $\delta$ now become two distinct analytical processes $(a_1, a_2)$ generated via the response-preference task (\Cref{eq:GRM}):
\begin{equation}
(a', p') \sim f_\theta(\ \cdot \mid q, r_1, r_2; a_1, a_2)
\label{eq:self-reflection}
\end{equation}
This allows the model to act as a meta-judge to evaluate its own analysis.
To avoid potential confusion between the model's input and output, we refer to the candidate analyses under evaluation as \textit{critiques}, as illustrated in the "Pairwise Analysis Preference" box in \Cref{figure:main_pipeline}.

\subsection{Training under the Unified Framework}
\label{ssec:training}
\paragraph{Data Collection} 
To develop the model's core reward modeling capability, we utilize standard \textit{Preference Data} (abbreviated as \textbf{Pref.}). Each instance consists of a user query $q$, a pair of candidate responses $(r_1, r_2)$, and a ground-truth label $y$ indicating the preferred response. Furthermore, we propose a novel data format termed \textit{Reflection Data} (abbreviated as \textbf{Refl.}) to supervise the quality of the analytical process. Similar to the Chain-of-Thought (CoT) process in reasoning models, the analytical trace in GRMs directly determines the reliability of its judgment. Building on this, we generate multiple outputs for preference data and pair their analyses based on the correctness of their predictions to construct Refl. data.

\paragraph{Unified Training} 
Based on the framework in \Cref{ssec:unified_framework}, we mix the Pref. data and Refl. data for unified training. Our unified framework ensures that these two types of data signals guide the same core preference judgment capability, rather than serving as unrelated objectives. This inherent connection allows us to combine both datasets into a single training process without causing task conflict, enabling ReflectRM to internalize a more robust and reliable evaluative logic.

It is important to note that the supervision signal for Refl. data does not require the model to imitate a perfect or "absolute" analytical process. Instead, it guides the model to identify which analysis is relatively more reasonable when comparing a correct trajectory with an incorrect one, mitigating the risk of rewarding post-hoc rationalizations.

\subsection{Two-Stage Inference via Self-Reflection}
\label{ssec:inference}
Following \Cref{ssec:training}, ReflectRM concurrently acquires response preference and self-reflection capabilities. We leverage these dual capabilities to perform a two-stage inference strategy, as illustrated in the bottom of \Cref{figure:main_pipeline}.


\paragraph{Stage 1: Rollouts and Confidence-guided Anchor Selection}

Given a query $q$ and response pair $(r_1, r_2)$, we perform $N$ independent rollouts (where $N=8$ in our experiments) using the response preference capability defined in \Cref{eq:GRM}. This produces an output set $\mathcal{O} =\{O_1, O_2, ..., O_N\}$, where each output $O_i$ is a tuple consisting of an analysis and a prediction:
\begin{equation}
O_i = (a_i, p_i), \quad p_i \in \{1, 2\}
\end{equation}
Specifically, the model's textual judgment is mapped to the index of the preferred response $\{1, 2\}$,
where $p_i=1$ indicates that $r_1$ is better than $r_2$, and $p_i=2$ indicates the opposite.

To effectively leverage the self-reflection capability, we select a high-confidence output from $\mathcal{O}$ to serve as a reliable anchor based on the generation probability \cite{fu2025deep}. For each output $O_i$, we identify the bottom 10\% of tokens with the lowest log-probabilities within the generated sequence, denoted as $T_{\text{bottom}}$, and calculate the confidence score as follows:

\begin{equation}
\text{Conf}(O_i) = \frac{1}{|T_{\text{bottom}}|} \sum_{t_j \in T_{\text{bottom}}} \log P(t_j)
\label{eq:confidence}
\end{equation}
The anchor output $O_{\text{anchor}}$ is then defined as the one with the highest confidence score:
\begin{equation}
O_{\text{anchor}} = \underset{O_i \in \mathcal{O}}{\text{argmax}}\ \text{Conf}(O_i)
\end{equation}
This method identifies the most internally coherent output to serve as a high-quality baseline from a response-level preference perspective.

%
\paragraph{Stage 2: Self-Reflection and Voting.}
Then, we use the self-reflection capability to evaluate whether other candidate analyses are better than the anchor, as formulated in \Cref{eq:self-reflection}. For each output $O_i$ ($i \neq \text{anchor}$),  we generate a reflection result $(a'_i, p'_i)$ by treating $a_i$ and $a_{\text{anchor}}$ as candidate analyses in a random order:
\begin{equation}
(a'_i, p'_i) \sim f_\theta(\ \cdot \mid q, r_1, r_2; a_i, a_{\text{anchor}})
\label{eq:inference_reflection}
\end{equation}
the result $p'_i$ reveals whether the model prefers analysis $a_i$ over $a_{\text{anchor}}$. We then collect all outputs that outperform the anchor into a \textit{Winner Group} $\mathcal{W}$:
\begin{equation}
\mathcal{W} = \{O_i \mid i \neq \text{anchor and } O_i \succ O_{\text{anchor}}\}
\label{eq:winner_group}
\end{equation}
where $O_i \succ O_{\text{anchor}}$ indicates that the reflection result $p'_i$ chooses $a_i$ as the better analysis.

The final prediction $P_{\text{final}}$ is determined by a majority vote \citep{mahan2024generative, wang2024self} among the outputs in $\mathcal{W}$, expressed as:
\begin{equation}
P_{\text{final}} = \underset{k \in \{1, 2\}}{\text{argmax}} \sum_{O_i \in \mathcal{W}} \mathbb{I}(p_i = k)
\label{eq:majority_vote}
\end{equation}
where $\mathbb{I}$ is the indicator function.
Crucially, if $\mathcal{W}$ is empty or \Cref{eq:majority_vote} results in a tie, we include the anchor output by setting $\mathcal{W} \leftarrow \mathcal{W} \cup \{O_{\text{anchor}}\}$, and repeat the majority voting process.

The effectiveness of this mechanism is further illustrated in \Cref{appendix:case_study}, where a qualitative case study demonstrates how self-reflection identifies and corrects analytical oversights during inference.



\section{Experiments}
\label{sec:experiment}

\subsection{Setup}
\label{ssec:setup}
\paragraph{Training Dataset}
We construct our training data using the HelpSteer3-Preference dataset \citep{wang2025helpsteer3}, a large-scale collection of open-ended tasks spanning diverse domains such as STEM, coding, and multilingual scenarios. To build the training set, we first exclude easy cases where the base model consistently selects the correct preferred response across all trials. 
From the remaining pool, we specifically leverage instances with inconsistent preferences to derive the reflection data (Refl.).
The final training set comprises a 4:1 mixture of Pref. data and Refl. data. More details on data statistics and the construction pipeline are provided in \Cref{appendix:data_construction}.
To further evaluate the generalizability of ReflectRM across different data distributions, we also apply our framework to the Skywork-Reward-V2 dataset \citep{liu2025skywork}. Detailed results regarding its performance and positional consistency are provided in \Cref{appendix:train_distribution}.

\paragraph{Benchmarks}


To ensure a robust evaluation, we assess \textbf{ReflectRM} on five established benchmarks: (1) \textbf{JudgeBench} \citep{tan2024judgebench}, which evaluates objective correctness in challenging real-world tasks such as coding and reasoning; (2) \textbf{RewardBench} \citep{lambert2025rewardbench}, a standard for measuring alignment with general human preferences across chat, safety, and reasoning; (3) \textbf{RM-Bench} \citep{liu2024rm}, designed to test the model's ability to distinguish core substance from stylistic distractions; (4) \textbf{RMB} \citep{zhou2024rmb}, a comprehensive suite covering 49 real-world task categories; and (5) \textbf{PPE-Preference} \citep{frick2024evaluate}, comprising 16k human-labeled pairs sourced from unfiltered user interactions.

\begin{table*}[!ht]
\centering
\small
\renewcommand{\arraystretch}{1.2}
\resizebox{\textwidth}{!}{%
\begin{tabular}{@{}cl cc cccc cc@{}} 
\toprule
\multirow{2}{*}{\textbf{ID}} & \multirow{2}{*}{\textbf{System}} & \multicolumn{2}{c}{\textbf{Source}} & \multirow{2}{*}{\textbf{RewardBench}} & \multirow{2}{*}{\textbf{RM-Bench}} & \multirow{2}{*}{\textbf{RMB}} & \multirow{2}{*}{\textbf{PPE-Preference}} & \multicolumn{2}{c}{\textbf{Overall}} \\
\cmidrule(lr){3-4} \cmidrule(lr){9-10}
 & & \textbf{Pref.} & \textbf{Refl.} & & & & & \textbf{AVG} & \textbf{$\Delta$ ($\uparrow$)} \\
\midrule
\multicolumn{10}{c}{\textit{Open Generative Reward Models}} \\ 
1 & Llama-3-OffsetBias-RM-8B & - & - & 89.0 & 71.3 & 57.8 & 59.2 & 69.3 & - \\
2 & ArmoRM-Llama3-8B-v0.1 & - & - & 90.4 & 69.3 & 64.6 & 60.6 & 71.2 & - \\
3 & Skywork-Reward-Llama-3.1-8B-v0.2 & - & - & 93.1 & 72.1 & 66.6 & 62.2 & 73.5 & - \\
4 & GRAM-R$^2$-3B & - & - & - & 84.6 & - & - & - & - \\
5 & DeepSeek-GRM-27B (MetaRM) & - & - & 90.4 & - & 70.3 & 67.2 & - & - \\
6 & RM-R1-DeepSeek-Distilled-Qwen-32B & - & - & 90.9 & 83.9 & 69.8 & - & - & - \\
\midrule

\multicolumn{10}{c}{\textit{Implemented Existing Method}} \\ 
7 & Qwen3-4B & - & - & 84.8 & 75.4 & 70.4 & 60.1 & 72.7 & - \\
8 & 4 + RFT & \greentext{$\checkmark$} & \redtext{$\times$} & 84.0 & 76.5 & 71.2 & 62.5 & 73.5 & \greentext{0.8} \\

\cdashline{1-10} 
\addlinespace[1pt]
\multicolumn{10}{c}{\textit{Our Method}} \\ 

9 & 4 + Unified Training & \greentext{$\checkmark$} & \greentext{$\checkmark$} & 85.3 & 76.5 & 75.3 & 63.9 & 75.2 & \greentext{2.5} \\
\rowcolor[HTML]{C0C0C0}
10 & 6 + Two-Stage Inference (ReflectRM) & \greentext{$\checkmark$} & \greentext{$\checkmark$} & \textbf{86.7} & \textbf{77.7} & \textbf{77.1} & \textbf{64.2} & \textbf{76.4} & \greentext{3.7} \\

\midrule 
\addlinespace[2pt]

\multicolumn{10}{c}{\textit{Implemented Existing Method}} \\ 
11 & Qwen3-8B & - & - & 85.0 & 78.5 & 73.9 & 62.9 & 75.1 & - \\
12 & 8 + RFT & \greentext{$\checkmark$} & \redtext{$\times$} & 86.4 & 77.4 & \textbf{76.0} & \textbf{64.5} & 76.1 & \greentext{1.0} \\
\cdashline{1-10} 
\addlinespace[1pt]
\multicolumn{10}{c}{\textit{Our Method}} \\ 
13 & 8 + Unified Training & \greentext{$\checkmark$} & \greentext{$\checkmark$} & 87.9 & 81.2 & 73.0 & 64.3 & 76.6 & \greentext{1.5} \\
\rowcolor[HTML]{C0C0C0}
14 & 10 + Two-Stage Inference (ReflectRM) & \greentext{$\checkmark$} & \greentext{$\checkmark$} & \textbf{89.2} & \textbf{82.7} & 73.3 & 64.0 & \textbf{77.3} & \greentext{2.2} \\
\bottomrule
\end{tabular}%
}
\caption{Main results on four RM benchmarks. Qwen3-4B and Qwen3-8B are chosen as the backbone model. The best results under each backbone are labeled using \textbf{bold font}.}
\label{tab:standard_performance}
\end{table*}
\paragraph{Baselines}
We compare ReflectRM against three categories of baselines: (1) \textbf{Base Model}, the original instructed LLMs without preference fine-tuning; (2) \textbf{RFT}, our primary baseline trained exclusively on the preference dataset using reinforcement fine-tuning; 
and (3) \textbf{Open Generative Reward Models}, which include several leading performance open-source models. Specifically, this category comprises: Llama-3-OffsetBias-RM-8B \citep{park2024offsetbias}, ArmoRM-Llama3-8B-v0.1 \citep{wang2024interpretable}, Skywork-Reward-Llama-3.1-8B-v0.2 \citep{liu2024skywork}, GRAM-R$^2$-3B \citep{wang2026gram}, DeepSeek-GRM-27B(MetaRM) \citep{liu2025inference}, and RM-R1-DeepSeek-Distilled-Qwen-32B \citep{chen2025rm}. Together, these baselines offer a broad context for evaluating the performance and robustness of our method.

\paragraph{Implementation Details} 
We fine-tune Qwen3-4B and Qwen3-8B \citep{yang2025qwen3} backbones using the Group Relative Policy Optimization (GRPO) algorithm \citep{shao2024deepseekmath}, which eliminates the need for a separate value model by estimating advantages within sampled groups. 
Training is conducted for 3 epochs with a batch size of 64, a learning rate of 1e-6, and a maximum generation length of 1024 tokens. For decoding, we used a sampling temperature of 1.0 to generate $N=8$ outputs for rollouts, and greedy decoding otherwise for deterministic evaluation. All experiments were conducted on 16 NVIDIA H800 GPUs.

\begin{figure}[!t]
    \centering
    \includegraphics[width=0.9\columnwidth]{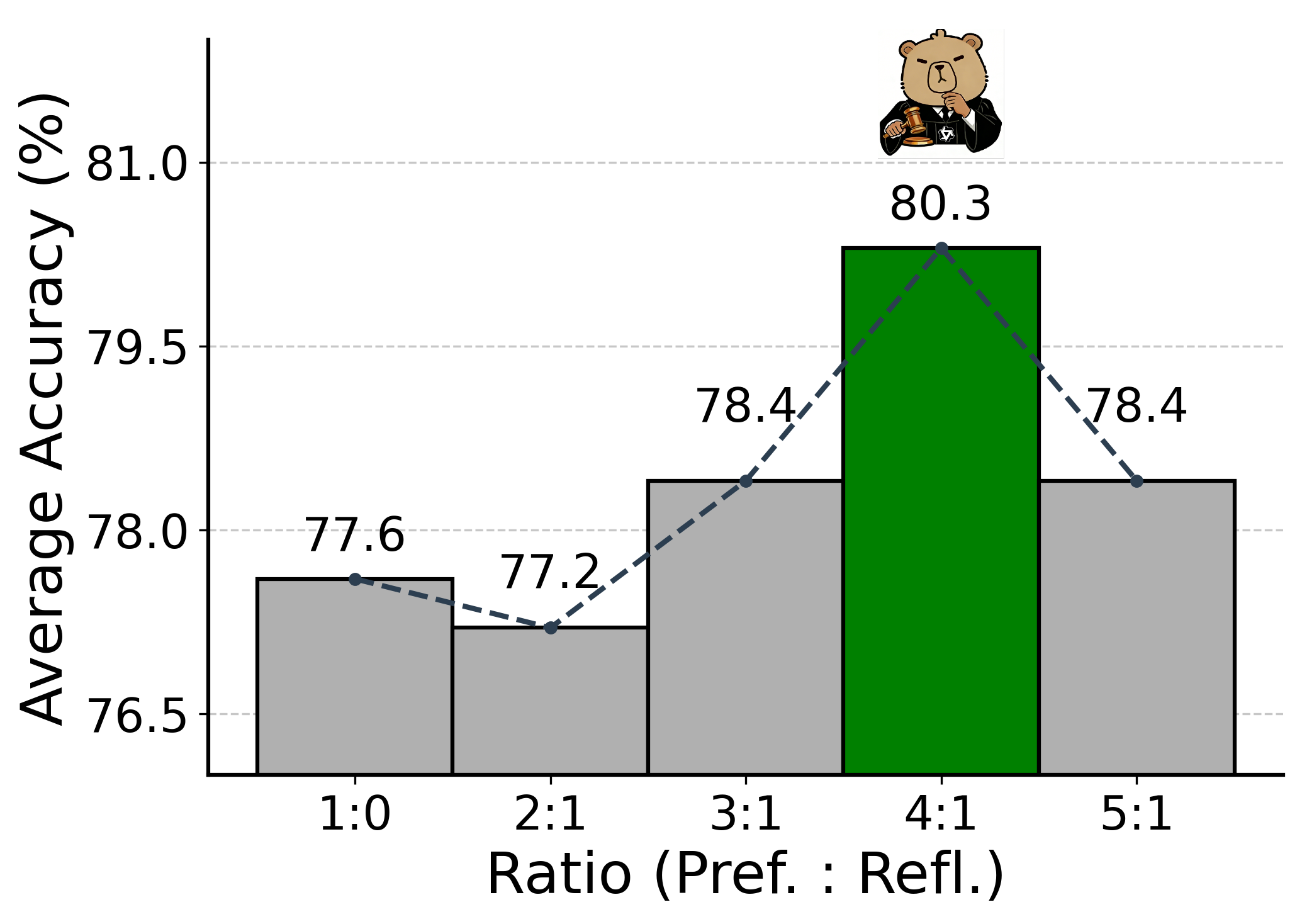} 
    \caption{Effect of preference-to-reflection ratio. A 4:1 ratio provides the optimal balance for training.} 
    \label{figure:ratio_analysis} 
\end{figure}

\subsection{Main Results}




The main performance results of ReflectRM are detailed in \Cref{tab:standard_performance}. The results show that while the standard RFT baseline achieves improvements over the base model, its performance is limited by the absence of process-level supervision (Systems 8 and 12). In contrast, incorporating reflection data into training yields significant performance gains across all model scales (Systems 9 and 13). These results confirm our hypothesis that, within our unified framework, learning to reflect on the analysis process and learning response preferences are mutually reinforcing. The reflection process provides richer supervisory signals, which in turn facilitate more effective learning of preference modeling.


As shown in \Cref{tab:standard_performance}, \textbf{ReflectRM} achieves the highest performance gains across different model scales, improving average accuracy by 3.7 points and 2.2 points for the 4B and 8B models, respectively. These improvements stem from the model's self-reflection capability, which identifies the most reliable analyses among multiple outputs to derive a more robust final prediction. Importantly, the gain is more pronounced for the smaller model: the 3.7-point improvement for Qwen3-4B is substantially larger than the 2.2-point gain for the 8B version. We hypothesize that this is because smaller-scale models exhibit more obvious analytical defects, which the self-reflection mechanism is particularly effective at identifying. Consequently, our enhanced smaller model even surpasses the larger baseline: System 10 attains an average accuracy of 76.4, outperforming System 12's score of 76.1.



\section{Analysis}
\label{sec:analysis}

This section aims to answer the following research questions: 
What are the optimal configurations for training and inference in ReflectRM? (\ref{ssec:optimal_config})
 How do the dual capabilities within our unified framework mutually reinforce each other? (\ref{ssec:mutual_reforcement})
How robust and scalable is ReflectRM across different settings? (\ref{ssec:robust_scalable})
Unless otherwise specified, all experiments are conducted using Qwen3-4B, with evaluations on RewardBench (abbreviated as R.B.) and RMB.

\subsection{Optimal Configurations of ReflectRM}
\label{ssec:optimal_config}

\begin{table}[!t]
\centering
\small
\resizebox{0.9\columnwidth}{!}{%
\begin{tabular}{l cccc} 
\toprule
\textbf{System} & \textbf{R.B.} & \textbf{RMB} & \textbf{AVG} & \textbf{$\Delta$} \\
\midrule
\textbf{ReflectRM} & 86.7 & 77.1 & 81.9 & - \\
\quad w/ Random Anchor  & 86.0 & 76.8 & 81.4 & \textcolor{red}{-0.5} \\
\quad w/ Random Winners & 85.7 & 76.0 & 80.9 & \textcolor{red}{-1.0} \\
\bottomrule
\end{tabular}%
}
\caption{Performance comparison on inference strategies. Both a reliable anchor and self-reflection capability are essential for ReflectRM.}
\label{tab:inference_strategy}
\end{table}


\begin{table}[!t]
\centering
\small
\resizebox{0.95\columnwidth}{!}{%
\begin{tabular}{l cc c  ccc }
\toprule
\textbf{System} & \textbf{Pref.} & \textbf{Refl.} & \textbf{Sum} & \textbf{R.B.} & \textbf{RMB} & \textbf{AVG} \\
\midrule
RFT & 13.7k & - & 13.7k & 84.0 & 71.2 & 77.6 \\
\quad w/ Scaled Pref. & 17.1k & - & 17.1k & 85.0 & 73.2 & 79.1 \\
\quad w/ Refl. (Ours) & 13.7k & 3.4k & 17.1k & \textbf{85.3} & \textbf{75.3} & \textbf{80.3} \\
\bottomrule
\end{tabular}%
}
\caption{Impact of reflection data on response preference. Reflection data offers richer learning signals compared to simply scaling preference data volume.}
\label{tab:reflection_effect}
\end{table}

\paragraph{Optimal Data Mixing Ratio}
While ReflectRM demonstrates strong performance in pairwise response preference judgment, identifying the optimal mixing ratio between \textbf{Pref.} and \textbf{Refl.} data remains a critical factor for our framework. We train several variants with ratios ranging from 1:0 (containing only Pref. data) to 5:1 (Pref. to Refl. data), all of which are evaluated using standard greedy decoding.
As illustrated in \Cref{figure:ratio_analysis}, the model's performance peaks at a 4:1 ratio. Interestingly, an excessive amount of reflection data (e.g., 2:1) leads to a slight decline in performance. We hypothesize that an excessive proportion of reflection tasks may distract the model from its primary objective of response preference modeling. These results confirm that a 4:1 ratio strikes an optimal balance for the unified training process.

\paragraph{Effectiveness of Inference Strategy}
To investigate the contribution of each component within our two-stage strategy, we evaluate its two key stages: confidence-guided anchor selection and self-reflection filtering. Specifically, we compare the full ReflectRM against two variants: (1) \textbf{Random Anchor}, which selects the anchor output randomly rather than via confidence scores; and (2) \textbf{Random Winners}, which replaces the self-reflection process with random sampling, selecting a number of outputs equal to the size of ReflectRM’s original winner group.

\begin{figure}[!t]
    \centering
    \includegraphics[width=0.9\columnwidth]{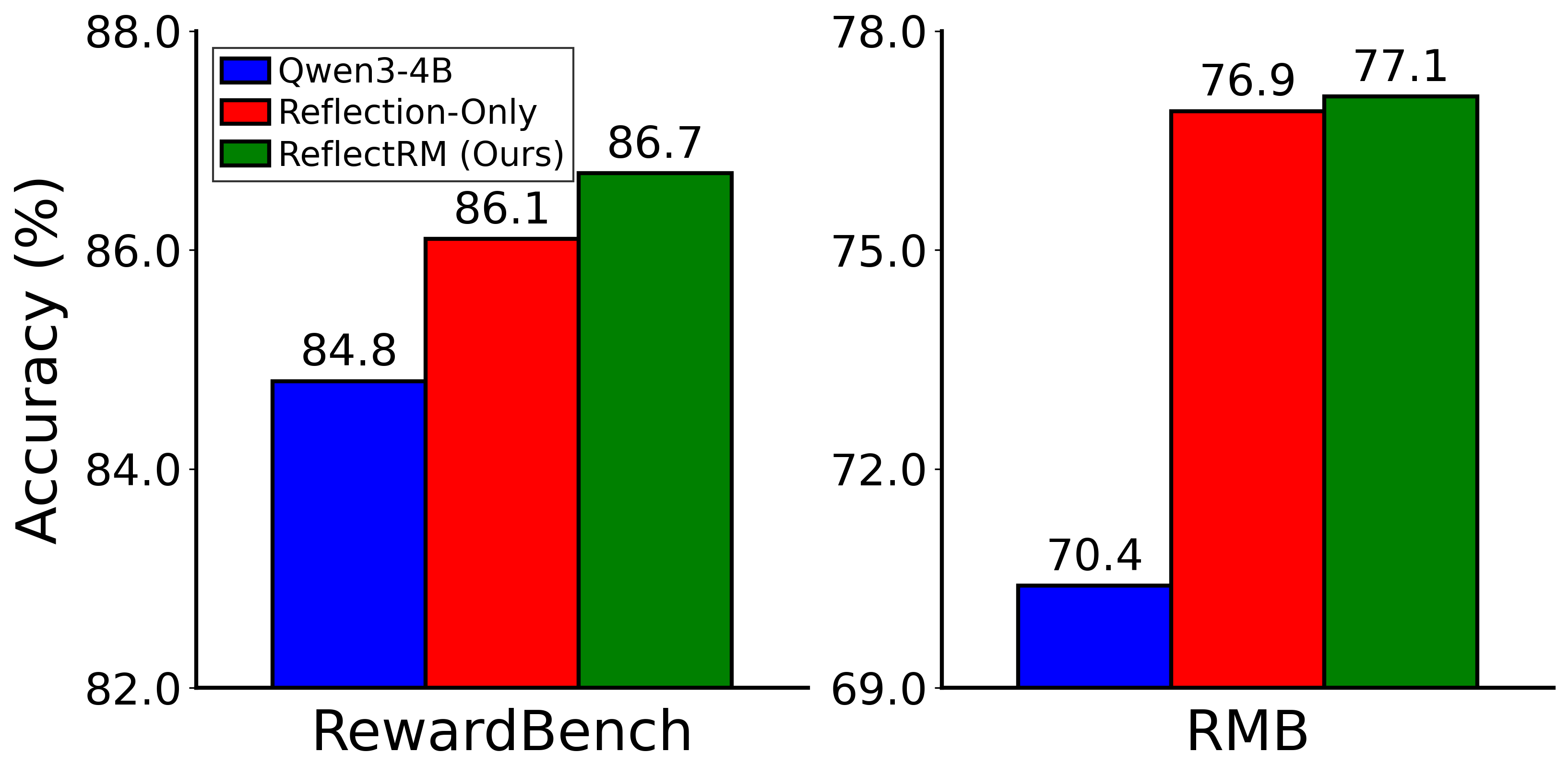}
    \caption{Impact of preference data on self-reflection. Preference data provides a robust foundation for the evaluative logic used to judge analytical processes.}
    \label{figure:decoupled_inference}
\end{figure}
\begin{table}[!t]
\centering
\small
\renewcommand{\arraystretch}{1.2}
\resizebox{0.85\columnwidth}{!}{%
\begin{tabular}{l ccc}
\toprule
\textbf{System} & \textbf{Qwen3-4B} & \textbf{Qwen3-8B} & \textbf{Qwen3-14B} \\
\midrule
Base Model & 72.7 & 75.1 & 75.7 \\
\textbf{ReflectRM (Ours)} & \textbf{76.4} & \textbf{77.3} & \textbf{79.5} \\
\bottomrule
\end{tabular}%
}
\caption{Performance scaling across model sizes. ReflectRM yields consistent performance gains, demonstrating the framework's high scalability.}
\label{tab:model_size}
\end{table}

\begin{table*}[!t]
\centering
\small
\renewcommand{\arraystretch}{1.2}
\resizebox{\textwidth}{!}{%
\begin{tabular}{@{}cl ccccc cc@{}} 
\toprule
\multirow{2}{*}{\textbf{ID}} & \multirow{2}{*}{\textbf{System}} & \multirow{2}{*}{\textbf{JudgeBench}} & \multirow{2}{*}{\textbf{RewardBench}} & \multirow{2}{*}{\textbf{RM-Bench}} & \multirow{2}{*}{\textbf{RMB}} & \multirow{2}{*}{\textbf{PPE-Preference}} & \multicolumn{2}{c}{\textbf{Overall}} \\
\cmidrule(lr){8-9}
 & & & (1k) & (1k) & (1k) & (1k) & \textbf{AVG} & \textbf{$\Delta$ ($\uparrow$)} \\
\midrule

7 & Qwen3-4B & 49.3 & 74.8 & 60.5 & 52.3 & 34.9 & 54.3 & - \\
8 & 4 + RFT & 53.1 & 75.1 & \textbf{67.1} & 56.5 & 46.6 & 59.7 & \greentext{5.4} \\
\cdashline{1-9} 
\addlinespace[1pt] 
9 & 4 + Unified Training & 53.1 & 78.5 & 64.3 & 64.3 & 48.1 & 61.7 & \greentext{7.4} \\
10 & 6 + Two-Stage Inference (ReflectRM) & \textbf{54.9} & \textbf{81.1} & 66.6 & \textbf{68.3} & \textbf{51.7} & \textbf{64.5} & \greentext{10.2} \\ 

\midrule 
\addlinespace[2pt] 

11 & Qwen3-8B & 52.7 & 78.4 & 64.9 & 60.9 & 42.1 & 59.8 & - \\
12 & 8 + RFT & 48.9 & 78.8 & 65.6 & \textbf{67.0} & \textbf{44.6} & 61.0 & \greentext{1.2} \\
\cdashline{1-9} 
\addlinespace[1pt] 
13 & 8 + Unified Training & 53.9 & 83.6 & 72.6 & 58.4 & 44.2 & 62.5 & \greentext{2.7} \\
14 & 10 + Two-Stage Inference (ReflectRM) & \textbf{56.3} & \textbf{84.9} & \textbf{73.5} & 59.4 & 44.3 & \textbf{63.7} & \greentext{3.9} \\ 
\bottomrule
\end{tabular}%
}
\caption{ Evaluation of positional consistency across five benchmarks. Qwen3-4B and Qwen3-8B are chosen as the backbone model. The best results under each backbone are labeled using \textbf{bold font}. ReflectRM achieves a substantial improvement of up to +10.2 points, nearly doubling the consistency gains of the standard RFT baseline, showing that supervising the analytical trace leads to a more stable and reliable judgment process.}
\label{tab:consistency_performance}
\end{table*}

As shown in \Cref{tab:inference_strategy}, both variants lead to performance degradation. Notably, the drop is significantly more pronounced when the self-reflection stage is bypassed, demonstrating that the model’s capability to evaluate its own analysis is the primary driver of inference-time performance gains. This validates that combining a high-confidence anchor with self-reflection filtering is essential for the effectiveness of ReflectRM. Besides, we provide a more comprehensive ablation in \Cref{appendix:ablation_on_infer}, including comparisons with simpler test-time strategies (e.g., Review \& Revise) and evaluations under matched computational budgets to isolate the source of our performance gains.



\subsection{Mutual Reinforcement in ReflectRM}
\label{ssec:mutual_reforcement}




\paragraph{Impact of Refl. on Response Preference}
We explore how reflection data benefits the model's core response preference capability. Specifically, we establish an initial baseline by training a model solely on the 13.7k Pref. data. To account for the influence of total training volume, we further construct a scaled variant using 17.1k samples of pure Pref. data. Finally, to ensure a fair comparison, all models in this experiment are evaluated using standard greedy decoding without utilizing the self-reflection capability at inference time. 


As shown in \Cref{tab:reflection_effect}, the model trained on mixed data significantly outperforms the variant trained on an equal volume of pure preference data. 
This result demonstrates that the reflection data provides a more effective learning signal, enhancing the model’s judgment capability more efficiently than simply scaling preference data alone. By explicitly supervising the analytical trace, the model internalizes a more robust underlying logic, which directly improves its response preference performance even without inference-time enhancements.
Notably, this performance gain is consistent across different model capacities. Extended experiments demonstrating the scalability of our mixed-data strategy are included in \Cref{appendix:scale_robust}.

\paragraph{Impact of Pref. on Analysis Preference}
Additionally, we investigate whether learning the response preference task provides an essential foundation for the self-reflection capability. To isolate this effect, we compare ReflectRM against a variant employing an independent reflection model (\textbf{Reflection-Only}, trained exclusively on Refl.) to perform the Inference Stage 2 described in \Cref{ssec:inference}. Crucially, both methods use the same initial outputs generated in Stage 1, ensuring the performance difference is driven solely by the quality of the self-reflection step. 

As illustrated in \Cref{figure:decoupled_inference}, ReflectRM consistently outperforms Reflection-Only. This gap illustrates that preference data also plays a vital role in enhancing the model's self-reflection capability. Within our unified framework, the ability to evaluate analytical processes is grounded in the core preference judgment ability learned from the preference data. As a result, ReflectRM can identify high-quality analytical traces with greater accuracy than a model trained only on reflection data. This provides strong evidence that the dual capabilities of ReflectRM are mutually reinforcing manifestations of the same underlying judgment ability, rather than a mere aggregation of disparate tasks.

\subsection{Robustness and Scalability of ReflectRM}
\label{ssec:robust_scalable}
To thoroughly investigate the robustness and scalability of ReflectRM, we conduct evaluations across all benchmarks. Specifically, scalability is reported based on four benchmarks: RewardBench, RM-Bench, RMB, and PPE-Preference. Positional bias is analyzed using the complete JudgeBench dataset and 1,000 randomly sampled instances from each of the other four benchmarks.
\paragraph{Scalability across Model Sizes}
We examine whether the performance gains of ReflectRM remain consistent as model capacity increases. We evaluate ReflectRM across 4B, 8B, and 14B parameter scales. As shown in \Cref{tab:model_size}, our method consistently yields improvements for all model sizes. These results demonstrate that ReflectRM is a scalable framework that provides a consistent boost independent of model capacity.

\paragraph{Robustness to Positional Bias}


Positional bias, the tendency for reward models to favor a response based on its presentation order (e.g., always preferring the first candidate) rather than its content, is a major challenge that leads to inconsistent and unstable judgments. In this part, we evaluate the effectiveness of ReflectRM in mitigating this bias by measuring positional consistency, where a sample is considered correct only if the model identifies the preferred response in \textbf{both} possible orderings. Specifically, we conduct thorough tests across five benchmarks while maintaining the same training configurations described in \Cref{sec:experiment}.


As shown in \Cref{tab:consistency_performance}, ReflectRM shows a significant improvement in positional consistency. Our method achieves a \textbf{10.2-point gain} over the base model, which is nearly double the 5.4-point gain of the RFT baseline.
Notably, this improvement in consistency (10.2) is far more substantial than the gain in standard accuracy (3.7) reported in \Cref{tab:standard_performance}. This disparity highlights the unique advantage of ReflectRM: by critiquing its own analytical process, the model learns to align its final decision with the analytical logic rather than positional orderings. This makes ReflectRM not only more accurate but also a far more reliable evaluator, demonstrating its significant potential to mitigate the long-standing challenge of positional bias in reward modeling.

\section{Conclusion}

We introduced \textbf{ReflectRM}, a framework that unifies response preference and process evaluation into a single generative objective. Our results demonstrate that these two capabilities are mutually reinforcing, enabling the model to internalize a more consistent and robust evaluative logic. By leveraging a two-stage inference strategy, ReflectRM significantly outperforms standard baselines across diverse benchmarks. Most notably, our method achieves a substantial improvement in positional consistency, demonstrating that supervising the analytical trace effectively mitigates positional bias and leads to more reliable reward modeling.

\section*{Limitations}
A primary limitation of ReflectRM is the increased computational overhead associated with its two-stage inference strategy. Compared to conventional inference-time scaling methods that perform majority voting across $N$ independent rollouts, our approach effectively doubles the computational requirements, as the self-reflection stage necessitates $N-1$ additional pairwise comparisons to filter the outputs. However, it’s worth noting that this extra computational cost can be mitigated through acceleration methods, such as quantification and speculative decoding.

\section*{Ethics Statement}
This work follows the ACL Ethics Policy. Our findings are based on publicly available datasets for reproducibility purposes. We acknowledge that LLMs may exhibit inherent societal biases and are prone to hallucinations. Therefore, if someone finds our work interesting and would like to use it in a specific environment, we strongly suggest the user conduct safety and bias evaluations to mitigate potential risks.

\section*{Acknowledgments}
This work was supported by the Shenzhen Science and Technology Program (Grant No. RCJC20210706091946001) and the Shenzhen Science and Technology Program (Grant No. ZDCY20250901104207008). We would like to thank the anonymous reviewers and meta-reviewer for their insightful suggestions.

\bibliography{main}

\clearpage
\appendix

\section{Appendix}
\label{sec:appendix}

\subsection{Details of Training Data Construction}
\label{appendix:data_construction}

The construction of the ReflectRM training dataset follows a two-step pipeline designed to generate high-quality supervision signals for both response preference modeling and process-level self-reflection.

\paragraph{Pref. Data} 
For each query in the HelpSteer3 dataset, we generate eight independent outputs using the instruction-tuned base model at a sampling temperature of 1.0. Each output consists of a textual analysis $a$ and a prediction $p$ indicating the preferred response. This sampling-based approach provides a diverse set of analytical paths for each problem. To focus on informative samples, we exclude easy instances where the model's prediction is correct across all eight trials, as these offer limited signals for further optimization. From the remaining pool, we randomly sample approximately 13.7k instances to form the \textbf{Pref.} data.

\paragraph{Refl. Data} 
To construct the reflection data, we leverage the subset of queries that yielded mixed outcomes (i.e., both correct and incorrect predictions) across the eight rollouts. For each such query, we pair the analysis from a correct output $a_{cor}$ with the analysis from an incorrect one $a_{inc}$ in a random order. Following the principle that the reliability of a GRM's final judgment is intrinsically determined by the quality of its analytical process, which is similar to the Chain-of-Thought (CoT) process in reasoning models, we designate ($a_{cor}$) as the preferred analysis. To maintain data diversity and balance, we generate exactly one reflection pair per query that exhibited inconsistent predictions. This automated pipeline allows us to synthesize thousands of analysis preference pairs without the need for manual annotation.

\paragraph{Final Training Dataset} 
As summarized in Table \ref{tab:data_statistics}, the final training set maintains a 4:1 ratio between standard preference data and reflection data. This mixture ratio was selected based on the empirical results in \Cref{ssec:optimal_config}, ensuring that the model maintains its primary capability in response preference judgment while effectively internalizing the self-reflection capability.

\begin{table}[!t]
\centering
\small
\resizebox{0.8\columnwidth}{!}{%
\begin{tabular}{lccc} 
\toprule
\textbf{Backbone} & \textbf{Pref.} & \textbf{Refl.} & \textbf{Sum} \\
\midrule
Qwen3-4B & 13,692 & 3,420 & 17,112 \\
Qwen3-8B & 13,560 & 3,380 & 16,940 \\
Qwen3-14B & 13,534 & 3,384 & 16,918 \\
\bottomrule
\end{tabular}%
}
\caption{Detailed statistics of the training dataset.}
\label{tab:data_statistics}
\end{table}

\subsection{Additional Ablations on Inference Mechanism}
\label{appendix:ablation_on_infer}
In this section, we provide a deeper investigation into the effectiveness of ReflectRM’s inference mechanism from two perspectives: strategy complexity and computational cost.

\subsubsection{Comparison with Simpler Test-time Strategies}
We further isolate the contribution of the self-reflection process by comparing ReflectRM against two simpler variants: (1) \textbf{Anchor-only}: Simply selecting the rollout with the highest confidence score as the final decision, bypassing the self-reflection mechanism; and (2) \textbf{Review \& Revise}: A standard two-pass refinement strategy where the model critiques and updates its initial judgment. 

\begin{table}[!t]
\centering
\small
\resizebox{0.95\columnwidth}{!}{%
\begin{tabular}{l cccc} 
\toprule
\textbf{System} & \textbf{R.B.} & \textbf{RMB} & \textbf{AVG} & \textbf{$\Delta$} \\
\midrule
ReflectRM & 86.7 & 77.1 & 81.9 & - \\
\quad w/ Anchor-only  & 81.1 & 74.5 & 77.8 & \textcolor{red}{-4.1} \\
Review \& Revise  & 84.5 & 71.5 & 78.0 & \textcolor{red}{-3.9} \\
\bottomrule
\end{tabular}%
}
\caption{Comparison with simpler inference strategies.}
\label{tab:ablation_on_infer}
\end{table}

The results in \Cref{tab:ablation_on_infer} show that ReflectRM outperforms both baselines. The gap between ReflectRM and "Anchor-only" confirms that while a high-confidence anchor is a strong starting point, the aggregation of multiple reliable analytical traces via reflection is crucial for robust judgment. Furthermore, the superiority over "Review \& Revise" suggests that our performance gains stem from the specialized self-reflection process rather than a mere second-chance mechanism.

\subsubsection{Performance under Matched Computational Budgets}
\begin{table*}[!t]
\centering
\small
\resizebox{0.9\textwidth}{!}{%
\begin{tabular}{l cc c  ccc }
\toprule
\textbf{System} & \textbf{RewardBench} & \textbf{RM-Bench} & \textbf{RMB} & \textbf{PPE-Preference} & \textbf{AVG} & \textbf{$\Delta$} \\
\midrule
GRM & 84.0 & 76.5 & 71.3 & 62.7 & 73.6 & - \\
ReflectRM (Ours) & 86.7 & 77.7 & 77.1 & 64.2 & 76.4 & \greentext{+2.8} \\
\bottomrule
\end{tabular}%
}
\caption{Performance comparison under matched computational budgets.}
\label{tab:ablation_on_budget}
\end{table*}

\begin{table*}[!t]
\centering
\small
\resizebox{0.9\textwidth}{!}{%
\begin{tabular}{l cc c  ccc }
\toprule
\textbf{System} & \textbf{RewardBench} & \textbf{RM-Bench} & \textbf{RMB} & \textbf{PPE-Preference} & \textbf{AVG} & \textbf{$\Delta$} \\
\midrule
Qwen3-4B & 84.8 & 75.4 & 70.4 & 60.1 & 72.7 & - \\
\quad + RFT & 83.1 & 75.1 & 74.1 & 62.2 & 73.6 & \greentext{+0.9} \\
\quad + ReflectRM (Ours) & 86.7 & 76.1 & 72.5 & 64.1 & 74.9 & \greentext{+2.2} \\
\bottomrule
\end{tabular}%
}
\caption{Main Performance on Skywork-Reward-V2 with Qwen3-4B.}
\label{tab:train_distribution_main}
\end{table*}

\begin{table*}[!t]
\centering
\small
\resizebox{\textwidth}{!}{%
\begin{tabular}{l ccc c  ccc }
\toprule
\textbf{System} & \textbf{JudgeBench} & \textbf{RewardBench (1k)} & \textbf{RM-Bench (1k)} & \textbf{RMB (1k)} & \textbf{PPE-Preference (1k)} & \textbf{AVG} & \textbf{$\Delta$} \\
\midrule
Qwen3-4B & 49.3 & 74.8 & 60.5 & 52.3 & 34.9 & 54.3 & - \\
\quad + RFT & 54.6 & 77.6 & 67.5 & 57.5 & 36.2 & 58.7 & \greentext{+4.4} \\
\quad + ReflectRM (Ours) & 57.6 & 79.6 & 69.1 & 59.3 & 40.8 & 61.3 & \greentext{+7.0} \\
\bottomrule
\end{tabular}%
}
\caption{Positional Consistency on Skywork-Reward-V2 with Qwen3-4B.}
\label{tab:train_distribution_consistency}
\end{table*}



\begin{table}[!t]
\centering
\small
\resizebox{0.9\columnwidth}{!}{%
\begin{tabular}{l cc c  ccc }
\toprule
\textbf{System} & \textbf{R.B.} & \textbf{RMB} & \textbf{AVG} \\
\midrule
Qwen3-4B & 84.8 & 70.4 & 77.6 \\
\quad + Pref. & 84.0 & 71.2 & 77.6 \\
\quad + Pref. \& Refl. & 85.3 & 75.3 & 80.3 \\
\cdashline{1-4} 
\addlinespace[2pt] 
Qwen3-8B & 85.8 & 72.9 & 79.4 \\
\quad + Pref. & 85.0 & 73.9 & 79.5 \\
\quad + Pref. \& Refl. & 87.9 & 73.0 & 80.5 \\
\cdashline{1-4} 
\addlinespace[2pt] 
Llama-3.1-8B-Instruct & 74.8 & 62.7 & 68.8 \\
\quad + Pref. & 75.6 & 63.9 & 69.8 \\
\quad + Pref. \& Refl. & 76.6 & 65.9 & 71.3 \\
\bottomrule
\end{tabular}%
}
\caption{Performance on RewardBench (R.B.) and RMB with mixed training data across different scales and architectures.}
\label{tab:different_model}
\end{table}
A potential concern is whether the gains of ReflectRM stem simply from increased computation during inference. To ensure a fair comparison, we evaluate a baseline GRM using Majority Voting with $2N-1 = 15$ rollouts, which matches the total computational budget of ReflectRM ($N = 8$ initial rollouts plus $N-1 = 7$ pairwise reflection comparisons).

As shown in \Cref{tab:ablation_on_budget}, even when the baseline is granted an equivalent computational budget, ReflectRM maintains a significant performance lead. This demonstrates that the gain of our method is derived from the structured evaluative logic of the self-reflection voting mechanism rather than a mere increase in inference-time compute.

\subsection{Robustness Across Different Training Distributions}
\label{appendix:train_distribution}
To demonstrate the robustness of ReflectRM across different training distributions, we train Qwen3-4B on Skywork-Reward-V2 dataset \citep{liu2025skywork} with the same 4:1 mixture ratio. 

As shown in \Cref{tab:train_distribution_main} and \Cref{tab:train_distribution_consistency}, ReflectRM yields consistent improvements in both overall accuracy and positional consistency compared to the RFT baseline. These results confirm that the benefits of process-level supervision are not dataset-specific but generalize effectively to other preference-training distributions.

\subsection{Scalability and Cross-Architecture Robustness}
\label{appendix:scale_robust}
We investigate whether the performance gains of ReflectRM scale with model capacity and generalize across architectures. We extend the experiments in \Cref{tab:reflection_effect} to include Qwen3-8B and Llama-3.1-8B-Instruct \cite{grattafiori2024llama}. 

As summarized in \Cref{tab:different_model}, the improvement remains consistent across different model sizes and families, confirming that our unified judgment framework is a scalable and robust solution for generative reward modeling.

\subsection{Prompt Templates}
\label{appendix:prompt_templates}

\begin{figure*}[t]
\begin{promptbox}{Prompt of Response Preference}
You are a discerning and impartial Judge. In the context of the conversation provided below (with the user's query being the last round), your role is to evaluate the quality of the two 'Responses' and determine which is better.\nYour decision should be based on which response better aligns with the user's instructions and more effectively addresses their query. To render a fair judgment, you need to think step-by-step to conduct a deep analysis, clearly articulating the reasoning for your decision.\nYour judgment must be free of any positional or length biases.

### Context
[The Begin of Conversation Context & Query]\n<context>\n[The End of Conversation Context & Query]

### Responses for Judgment
These are the two responses you must analyze and compare.

[The Begin of Response 1]\n<response 1>\n[The End of Response 1]
[The Begin of Response 2]\n<response 2>\n[The End of Response 2]

---

### Your Structured Judgment
Follow these steps precisely and use the specified tags for your output.\n**1. Provide Detailed Analysis:** Think step-by-step to conduct a detailed analysis of the Responses. Place your analysis within `<Analysis>` tags.\n**2. Render Final Verdict:** Conclude with your final verdict based on your analysis. State which response is better in the format "Response 1 is better than Response 2" or "Response 2 is better than Response 1". Place this final choice within `<Result>` tags.\nYour entire output must follow the format below.\n<Analysis>\nYour detailed step-by-step analysis of the two responses.\n</Analysis>\n<Result>\nBased on your Analysis, only print the following: "Response 1 is better than Response 2" OR "Response 2 is better than Response 1".\n</Result>/no_think
\end{promptbox}
\vspace{-0.5em} 

\vspace{2em} 

\begin{promptbox}{Prompt of Analysis Preference}
You are a discerning and impartial Judge. Your role is to evaluate the quality of the two 'Critiques' presented below. These critiques are themselves analyses of two original 'Responses' generated for a conversation (with the user's query being the last round).\nYour decision should be based on which critique provides a more insightful, accurate, fair, and well-reasoned analysis. To render a fair judgment, you need to think step-by-step to conduct a deep analysis, clearly articulating the reasoning for your decision.\nYour judgment must be free of any positional or length biases.

### Context
[The Begin of Conversation Context & Query]\n<context>\n[The End of Conversation Context & Query]
[The Begin of Response 1]\n<response 1>\n[The End of Response 1]
[The Begin of Response 2]\n<response 2>\n[The End of Response 2]

### Critiques for Judgment
These are the two critiques you must analyze and compare. Each one analyzes 'Response 1' and 'Response 2' shown in 'Context'.

[The Begin of Critique 1]\n<critique 1>\n[The End of Critique 1]
[The Begin of Critique 2]\n<critique 2>\n[The End of Critique 2]

---

### Your Structured Judgment
Follow these steps precisely and use the specified tags for your output.\n**1. Provide Detailed Analysis:** Think step-by-step to conduct a detailed analysis of the Critiques. Place your analysis within `<Analysis>` tags.\n**2. Render Final Verdict:** Conclude with your final verdict based on your analysis. State which critique is better in the format "Critique 1 is better than Critique 2" or "Critique 2 is better than Critique 1". Place this final choice within `<Result>` tags.\nYour entire output must follow the format below.\n<Analysis>\nYour detailed step-by-step analysis of the two critiques.\n</Analysis>\n<Result>\nBased on your Analysis, only print the following: "Critique 1 is better than Critique 2" OR "Critique 2 is better than Critique 1".\n</Result>/no_think
\end{promptbox}
\vspace{-0.5em}
\caption{Prompt templates for the unified judgment framework. The top template is used for pairwise response preference judgment, while the bottom one is for pairwise analysis preference (self-reflection).}
\label{fig:prompt_templates}
\end{figure*}

Following our unified framework described in \Cref{ssec:unified_framework}, we implement a standardized prompting strategy to accommodate response preference and analysis preference tasks. As illustrated in \Cref{fig:prompt_templates}, the templates are designed to align with the components of \Cref{eq:unified_perspective}:
\begin{itemize}
    \item \textbf{Condition $\phi$}: This is represented by the \texttt{Context} field. For response preference, $\phi$ includes the conversation history and the user's query $q$. For analysis preference, $\phi$ is expanded to include the two candidate responses $(r_1, r_2)$ to provide the necessary context for evaluating the subsequent analyses (i.e. critiques).
    \item \textbf{Candidates $\delta$}: This part corresponds to the \texttt{Responses for Judgment} or \texttt{Critiques for Judgment} fields, containing the two entities to be compared.
    \item \textbf{Output $(a,p)$}: We mandate a structured output format using \texttt{<Analysis>} and \texttt{<Result>} tags. The content within \texttt{<Analysis>} corresponds to the textual analysis $a$. The \texttt{<Result>} tag contains a preference statement (e.g., Response 1 is better than Response 2), which is mapped to the preference index $p \in \{1,2\}$.
\end{itemize}

Notably, we append the \texttt{</no\_think>} token at the end of each prompt. This is a model-specific control token for the Qwen3 backbone that deactivates its internal reasoning mode.
This prompting design ensures strict alignment with our unified generative formulation while maintaining clear task distinction, effectively preventing the model from confusing the two preference capabilities.


\subsection{Case Study}
\label{appendix:case_study}
\begin{table*}[t]
    \centering
    \small 
    \renewcommand{\arraystretch}{1.6} 
    \begin{tabularx}{\textwidth}{L X}
        \toprule
        \rowcolor{gray!15} 
        \multicolumn{2}{c}{\textbf{Query and Responses}} \\
        
        \multicolumn{1}{c}{\textbf{Query}} & In this conversation, please keep your responses succint the to the point. Please avoid lengthy explanations or examples unless instructed. \\
        \multicolumn{1}{c}{\textbf{Response 1}} & Understood! I'll keep my responses brief and to the point. \\
        \multicolumn{1}{c}{\textbf{Response 2}} & Absolutely, I'll keep my responses concise and to the point. How can I assist you today? \\
        
        \midrule \midrule
        \rowcolor{gray!15}
        \multicolumn{2}{c}{\textbf{Output 1}} \\
        
         \multicolumn{1}{c}{\textbf{Analysis}} & The user requested that responses be ``succint the to the point'' and to avoid lengthy explanations or examples unless instructed. Response 1 simply says ``Understood! I'll keep my responses brief and to the point.'' while Response 2 adds ``Absolutely, I'll keep my responses concise and to the point. How can I assist you today?'' \par \smallskip
        Response 2 is more comprehensive as it not only confirms understanding of the user's instruction but also actively asks how the user wants to proceed, which shows engagement and willingness to assist. However, the user's query was simply to confirm that the responses would be brief and to the point, and the initial response from the assistant was already meeting that requirement. Response 1 is more in line with the user's explicit instruction to be ``succint the to the point'' as it directly confirms understanding without adding extra information. Response 2, while polite, includes an additional question that \colorbox{red!20}{may not be necessary} if the user's query was solely about confirming the response style. \colorbox{yellow!50}{\texttt{Critique 1}} \colorbox{blue!20}{\texttt{Anchor}} \\
        
        \multicolumn{1}{c}{\textbf{Predict}} & \textit{Response 1 is better than Response 2} \redtext{$\times$} \\
        
        \midrule
        \rowcolor{gray!15}
        \multicolumn{2}{c}{\textbf{Output 2}} \\
        
        \multicolumn{1}{c}{\textbf{Analysis}} & The user instructed to keep responses succinct and to the point, avoiding lengthy explanations or examples unless instructed. Response 1 simply says ``Understood! I'll keep my responses brief and to the point.'' This directly acknowledges the instruction but does not offer any further assistance or information. Response 2, however, not only acknowledges the instruction but also actively asks how the user wants to proceed, showing a willingness to assist and engage in the conversation. While both responses are concise, Response 2 provides a more effective alignment with the user's instructions by demonstrating proactive engagement and offering further assistance, \colorbox{green!20}{which is more in line with the context of a conversational interaction}. \colorbox{yellow!50}{\texttt{Critique 2}} \\
        
        \multicolumn{1}{c}{\textbf{Predict}} & \textit{Response 2 is better than Response 1} \greentext{$\checkmark$} \\
        
        \midrule
        \rowcolor{gray!15}
        \multicolumn{2}{c}{\textbf{Self-Reflection}} \\
        
        \multicolumn{1}{c}{\textbf{Analysis}} & The two critiques provide different perspectives on the analysis of Response 1 and Response 2. Critique 1 focuses on the difference between the two responses by highlighting that Response 2 includes an additional question that may not be necessary if the user's query was solely about confirming the response style. It argues that Response 1 is more in line with the user's explicit instruction because it directly confirms understanding without adding extra information. However, Critique 2 emphasizes that Response 2 provides a more effective alignment with the user's instructions by demonstrating proactive engagement and offering further assistance, which is more in line with the context of a conversational interaction. \par \smallskip
        Critique 2 offers a more comprehensive analysis by recognizing that while both responses are concise, Response 2 demonstrates proactive engagement and offers further assistance, which is crucial in a conversational context. It also acknowledges the user's instruction to keep responses succinct but emphasizes the importance of active engagement in a conversation. In contrast, Critique 1 is slightly more focused on the literal interpretation of the user's instruction and \colorbox{green!20}{may overlook the conversational context} by suggesting that the additional question in Response 2 is unnecessary. \\
        
        \multicolumn{1}{c}{\textbf{Predict}} & \textit{Critique 2 is better than Critique 1} \\
        
        \midrule\midrule
        \multicolumn{1}{c}{\textbf{Final Predict}} & \textbf{Response 2 is better than Response 1} \greentext{$\checkmark$} \\
        \bottomrule
    \end{tabularx}
    \caption{A case study of ReflectRM's two-stage inference strategy. By reflecting on the analytical processes, ReflectRM finally derives the correct verdict.}
    \label{tab:case_study}
\end{table*}
We present a case study of ReflectRM's two-stage inference pipeline in Table \ref{tab:case_study}. In this scenario, the user provides a specific instruction to keep responses "succinct and to the point". 
Notably, the original query from PPE-Preference contains minor errors (e.g., 'succint the to the point'). These are preserved verbatim to ensure the case study faithfully reflects the source benchmark.
Both responses are concise, but Response 2 includes a proactive follow-up question.

During the first stage of inference, the model generates $N=8$ outputs to judge the preference between two responses. We select two representative outputs to illustrate the process (including the anchor output). Interestingly, the selected anchor over-emphasizes the literal interpretation of the brevity constraint, penalizing Response 2's follow-up as "not necessary". Consequently, this high-confidence anchor initially leads to an incorrect preference prediction.

In the second stage, ReflectRM leverages its self-reflection capability to evaluate the analytical quality of outputs against the anchor. As shown in the \texttt{Self-Reflection} section of \Cref{tab:case_study}, the model identifies the analytical oversight in Critique 1, recognizing that Critique 2 provides a more comprehensive and context-aware evaluation, and finally derives a correct prediction.

This case clearly illustrates how the self-reflection capability allows ReflectRM to derive a more robust and reliable result.



\end{document}